%% file: main.tex
\documentclass[runningheads]{llncs}
\usepackage[T1]{fontenc}
\usepackage{graphicx}
\usepackage{booktabs}
\usepackage{amsmath}
\usepackage{amssymb}
\usepackage{multirow}
\usepackage{algorithm}
\usepackage{algorithmic}
\usepackage{hyperref}
\usepackage{bbm}
\usepackage{tikz}
\usepackage{subcaption}

\begin{document}

\title{When Can We Trust LLM Graders? Calibrating Confidence for Automated Assessment}
\titlerunning{Calibrating LLM Confidence for Automated Assessment}

\author{Robinson Vasquez Ferrer\and
Damla Turgut\and \\
Zhongzhou Chen\and
Shashank Sonkar} 
\institute{University of Central Florida \\
\email{\{robinson.vasquezferrer, shashank.sonkar\}@ucf.edu}}

\authorrunning{R. Ferrer et al.}

\maketitle

\input{sections/abstract}
\input{sections/intro}
\input{sections/related_work}
\input{sections/methodology}
\input{sections/experiments_results}
\input{sections/conclusion}

\end{document}

%% file: sections/abstract.tex
\begin{abstract}
Large Language Models (LLMs) show promise for automated grading, but their outputs can be unreliable. Rather than improving grading accuracy directly, we address a complementary problem: \textit{predicting when an LLM grader is likely to be correct}. This enables selective automation where high-confidence predictions are processed automatically while uncertain cases are flagged for human review. We compare three confidence estimation methods (self-reported confidence, self-consistency voting, and token probability) across seven LLMs of varying scale (4B to 120B parameters) on three educational datasets: RiceChem (long-answer chemistry), SciEntsBank, and Beetle (short-answer science). Our experiments reveal that self-reported confidence consistently achieves the best calibration across all conditions (avg ECE 0.166 vs 0.229 for self-consistency). Surprisingly, self-consistency remains 38\% worse despite requiring 5$\times$ the inference cost. Larger models exhibit substantially better calibration though gains vary by dataset and method (e.g., a 28\% ECE reduction for self-reported), with GPT-OSS-120B achieving the best calibration (avg ECE 0.100) and strong discrimination (avg AUC 0.668). We also observe that confidence is strongly top-skewed across methods, creating a ``confidence floor'' that practitioners must account for when setting thresholds. These findings suggest that simply asking LLMs to report their confidence provides a practical approach for identifying reliable grading predictions. Code is available \href{https://github.com/sonkar-lab/llm_grading_calibration}{here}.
\end{abstract}

\keywords{Large Language Models, Automated Grading, Confidence Calibration, Automated Assessment, Human-AI Collaboration, Short Answer Grading, Long Answer Grading}

%% file: sections/intro.tex
\section{Introduction}

Large Language Models (LLMs) are increasingly used for automated grading, but their reliability remains questionable. Studies have found that GPT-4 produces scores that differ from human evaluators~\cite{Kostic2024} and makes common arithmetic errors that reduce accuracy to 20\% on complex problems~\cite{Gandolfi2025}. While LLMs can match human agreement on some dimensions, they achieve only fair agreement on others~\cite{Tian2024Artifacts}, and performance degrades substantially for low-quality student work~\cite{Usher2025PeerGrading}. These limitations raise a fundamental question: \textit{when can we trust an LLM's grading decision?}

Rather than attempting to improve grading accuracy directly, we address a complementary problem: \textit{predicting when an LLM grader is likely to be correct}. This framing enables selective automation, where high-confidence predictions are processed automatically while uncertain cases are flagged for human review. Such human-AI collaboration can preserve grading quality while reducing instructor workload on the cases where LLMs are reliable.

The key to effective selective automation is \textit{confidence calibration}: a confidence score of 0.8 should correspond to 80\% accuracy among predictions at that confidence level. Well-calibrated confidence enables principled decisions about when to trust automated grading. Unfortunately, modern neural networks tend to be poorly calibrated, often exhibiting systematic overconfidence~\cite{Guo2017ICML}, and this problem persists in LLMs~\cite{Xiong2024ICLR}. Recent surveys have catalogued various approaches to confidence estimation and calibration in LLMs~\cite{Geng2024NAACL}.

We systematically evaluate three approaches to confidence estimation, each grounded in foundational NLP research: sampling-based consistency checks (self-consistency)~\cite{Wang2023SelfConsist}, prompting the model to verbalize its uncertainty (self-reported confidence)~\cite{Kadavath2022}, and leveraging internal probability mass over ``yes''/``no'' verdict tokens (token probability)~\cite{Kuhn2023ICLR}. 

We evaluate these methods across seven LLMs spanning two orders of magnitude in scale (4B to 120B parameters): Llama-8B/70B~\cite{MetaLlama3}, Qwen-4B/80B-A3B variants~\cite{yang2025qwen3}, and GPT-OSS-20B/120B~\cite{agarwal2025gpt}. We test on three established educational datasets: RiceChem for long-answer chemistry grading~\cite{RiceChem2024}, and SciEntsBank and Beetle for short-answer science assessment~\cite{Dzikovska2013SemEval}. Our primary calibration metrics are Expected Calibration Error (ECE) and the Brier score, which together measure how well confidence scores predict accuracy.

Our experiments yield several findings with practical implications:

\begin{itemize}
\item \textbf{Self-reported confidence achieves the best calibration.} Across all six evaluated conditions, simply prompting the models to state their confidence outperforms sampling-based alternatives. Despite requiring higher inference cost, self-consistency performs worse than self-reported.

\item \textbf{Model scale boosts accuracy more than calibration.} Larger models are consistently more accurate across all three datasets, while calibration gains are mixed and dataset/method-dependent. We report detailed per-dataset calibration changes and per-model comparisons in Results.

\item \textbf{Confidence is top-skewed across methods.} Confidence distributions are heavily concentrated at high values; self-consistency is discrete by construction, while self-reported and token-probability are continuous. These distributions create practical confidence floors, so thresholds should be set relative to each model/method’s observed scores rather than an intuitive midpoint.
\end{itemize}

These findings suggest that simply asking LLMs to report their confidence provides a practical approach for identifying reliable grading predictions. With appropriate threshold calibration relative to the model's confidence distribution, selective automation can process high-confidence predictions automatically while flagging uncertain cases for human review.

%% file: sections/related_work.tex
\section{Related Work}

\subsection{Calibration of Neural Networks}

Calibration, the correspondence between predicted confidence and actual accuracy, has been extensively studied in the machine learning literature. Early work established foundational calibration methods: \cite{Platt1999} introduced Platt scaling using sigmoid functions to calibrate SVM outputs, while \cite{ZadroznyElkan2002} proposed isotonic regression as a non-parametric alternative. \cite{Guo2017ICML} demonstrated that modern deep neural networks are poorly calibrated, typically exhibiting overconfidence, and proposed temperature scaling as a simple post-hoc method. \cite{Hendrycks2017} showed that maximum softmax probability serves as a baseline for detecting misclassified examples, though it remains imperfect. For language models specifically, \cite{Tian2023EMNLP} demonstrated that RLHF-trained models like GPT-4 produce verbalized confidence that is often better calibrated than internal probabilities. In this work, we focus on eliciting confidence from LLMs rather than applying post-hoc calibration.

\subsection{Confidence Estimation in NLP}

Several approaches have been proposed for estimating uncertainty in LLM outputs. \cite{Wang2023SelfConsist} introduced self-consistency, which samples multiple reasoning paths and takes a majority vote; the agreement level provides a natural confidence measure. \cite{Kuhn2023ICLR} proposed semantic entropy, which clusters sampled responses by meaning before computing uncertainty, providing a principled measure that is invariant to paraphrasing.

Ensemble-based methods offer another paradigm: \cite{Lakshminarayanan2017NeurIPS} showed that deep ensembles provide well-calibrated uncertainty estimates, while \cite{Gal2016ICML} demonstrated that MC Dropout approximates Bayesian inference for uncertainty estimation. However, these methods require multiple forward passes or model copies, which is prohibitive for large language models.

Self-reported confidence, explicitly prompting models to state their confidence, has emerged as a lightweight alternative. \cite{Kadavath2022} demonstrated that language models ``mostly know what they know,'' exhibiting metacognitive capabilities.

\subsection{Automated Short Answer Grading}

Automated Short Answer Grading (ASAG) has a rich history in educational data mining. The SciEntsBank and Beetle datasets, introduced by \cite{Dzikovska2012} and formalized as a shared task~\cite{Dzikovska2013SemEval}, have served as standard benchmarks. \cite{Burrows2015IJAIED} provided a comprehensive survey identifying 35 ASAG systems across five temporal eras.

The emergence of LLMs has transformed ASAG research. \cite{Jiang2024LAS} showed that GPT-4 can achieve substantial agreement with human graders (quadratic weighted kappa of 0.68) without task-specific training. For long-answer grading, the RiceChem dataset formulates the task as rubric entailment~\cite{RiceChem2024}.

While this prior work focuses on improving grading accuracy, our work addresses the complementary problem of predicting when automated grades are reliable. This confidence-aware perspective enables practical deployment with appropriate human oversight.

%% file: sections/methodology.tex
\section{Methodology}
\label{sec:methodology}

We formalize the grading task and then describe three methods for estimating confidence in LLM grading predictions.
We consistently prompt the model to emit a JSON dictionary whose `"verdict"` key holds the yes/no grading decision; all confidence methods operate on that verdict plus derived uncertainty signals.\footnote{Prompt templates, token sets, and code will be released once the codebase is organized for reproducibility.}

\subsection{Task Definition}

Let the binary grading task be defined as:
\begin{equation}
    T: (Q, S, R) \rightarrow \{0, 1\}
\end{equation}
where $Q$ is the question, $S$ is the student's response, and $R$ is a rubric item specifying what the response should demonstrate. The output is 1 if the student response satisfies the rubric item, 0 otherwise.

An LLM grader $M$ attempts to approximate $T$:
\begin{equation}
    M: (Q, S, R) \rightarrow \{0, 1\}
\end{equation}

However, $M(x) \neq T(x)$ for some inputs $x = (Q, S, R)$ due to LLM limitations. Our goal is to design a confidence estimator $C: (Q, S, R) \rightarrow [0, 1]$ such that $C(x)$ estimates:
\begin{equation}
    P(M(x) = T(x) \mid x)
\end{equation}

\subsection{Confidence Estimation Methods}

We compare three approaches that differ in computational cost and the type of uncertainty they capture.

\subsubsection{Token Probability}

This method directly uses the model's output probability distribution over verdict tokens.

For a binary grading prompt, we take the top-20 tokens by probability mass and aggregate the probabilities of case-insensitive yes/no variants within that set. Let $\mathcal{T}_{\text{yes}}$ and $\mathcal{T}_{\text{no}}$ denote the top-ranked token sets, and define
\begin{equation}
    P_{\text{yes}} = \sum_{t \in \mathcal{T}_{\text{yes}}} \text{softmax}(\text{logits})[t], \quad P_{\text{no}} = \sum_{t \in \mathcal{T}_{\text{no}}} \text{softmax}(\text{logits})[t]
\end{equation}

The prediction is determined by the token the LLM actually emitted for the JSON ``verdict'' field: $\hat{y}=1$ when that token is exactly ``yes'' (case-insensitive) and $\hat{y}=0$ when it is ``no''; we aggregate the probability mass over all yes/no variants only for the confidence, i.e.:
\begin{equation}
    C_{\text{token}}(x) =
    \begin{cases}
        P_{\text{yes}} & \text{if the verdict token is ``yes'' (case-insensitive)}\\
        P_{\text{no}} & \text{if the verdict token is ``no'' (case-insensitive)}
    \end{cases}
\end{equation}

\textbf{Cost:} 1 LLM call.

\subsubsection{Self-Reported Confidence}

This method prompts the model to explicitly report its confidence alongside the grading decision.

We prompt the model to emit both a verdict and a numeric confidence score, capturing its self-assessed uncertainty.

From the parsed output, we extract the binary decision $\hat{y} \in \{0, 1\}$ and raw confidence $c_{\text{raw}} \in \{1, 2, \ldots, 10\}$. The normalized confidence is:
\begin{equation}
    C_{\text{self}}(x) = \frac{c_{\text{raw}} - 1}{9}
\end{equation}

\textbf{Cost:} 1 LLM call. \textbf{Captures:} Model's self-assessed uncertainty based on internal reasoning.

\subsubsection{Self-Consistency}

This method measures agreement across multiple stochastic samples from the model.

Sample $n$ outputs from $M$ with temperature $\tau > 0$:
\begin{equation}
    \{y_1, y_2, \ldots, y_n\} \sim M(x, \tau)
\end{equation}

The self-consistency confidence is the majority vote fraction:
\begin{equation}
    C_{\text{SC}}(x) = \max\left(\frac{\sum_{i=1}^{n} \mathbbm{1}\{y_i = 0\}}{n}, \frac{\sum_{i=1}^{n} \mathbbm{1}\{y_i = 1\}}{n}\right)
\end{equation}

The final prediction is $\hat{y} = \mathbbm{1}\{\sum_{i=1}^{n} y_i > n/2\}$.

\textbf{Cost:} $n$ LLM calls (we use $n=5$).

\subsection{Evaluation Metrics}

\subsubsection{Expected Calibration Error (ECE)}\cite{Pakdaman_2015}

ECE measures the average difference between confidence and accuracy across confidence bins:
\begin{equation}
    \text{ECE} = \sum_{b=1}^{B} \frac{|B_b|}{n} \left| \text{acc}(B_b) - \text{conf}(B_b) \right|
\end{equation}
where $B_b$ is the set of predictions in bin $b$, $\text{acc}(B_b)$ is the accuracy within the bin, and $\text{conf}(B_b)$ is the mean confidence. Lower ECE indicates better calibration. We use $B=10$ equal-width bins.

\subsubsection{Brier Score}\cite{Brier1950}

The Brier score measures mean squared error between confidence and binary correctness:
\begin{equation}
    \text{BS} = \frac{1}{n} \sum_{i=1}^{n} (C(x_i) - \mathbbm{1}\{M(x_i) = T(x_i)\})^2
\end{equation}
Lower Brier scores indicate better calibration and discrimination combined.

\subsubsection{AUC-ROC}

The Area Under the ROC Curve measures ranking quality---whether higher confidence corresponds to higher accuracy. AUC-ROC ranges from 0.5 (random) to 1.0 (perfect ranking).~\cite{Fawcett2006}

\subsubsection{Coverage at Risk}

For practical deployment, we report coverage at fixed risk thresholds: the fraction of predictions that can be automated while maintaining accuracy above a target level.

\subsubsection{Statistical Methods}

We report 95\% confidence intervals for all primary metrics using a student-block bootstrap (1{,}000 draws, percentile intervals). Each bootstrap sample resamples students with replacement and keeps all of that student’s responses together, preserving within-student dependence.\cite{FieldWelsh2007,EfronTibshirani1993} To assess feature--calibration interactions, we fit linear mixed-effects models with random intercepts for students, answers and rubrics, and report standardized coefficients ($\beta$). Question effects are modeled as fixed effects for small question sets ($\leq20$) and as random intercept otherwise. \cite{fielding2006cross,Noortgate2003,Dalton2013} We extract features from rubrics, student answers, LLM responses, and question text. We also run paired bootstrap comparisons on identical instances to test method/model dominance.\cite{Koehn2004} We control false discovery rate using the Benjamini--Hochberg procedure.\cite{FDR_BH}

\noindent\textbf{Interpretation.} We treat claims supported by paired comparisons and confidence intervals as inferential (expected to generalize to similar ASAG tasks with comparable prompts, rubrics, and response distributions).

%% file: sections/experiments_results.tex
\section{Experiments and Results}

\subsection{Experimental Setup}

\textbf{Datasets.} We evaluate on three educational datasets covering different domains and answer lengths. \textbf{RiceChem}~\cite{RiceChem2024} contains long-answer chemistry questions requiring explanations, with 8,392 student responses across 4 questions and 27 rubric items. \textbf{SciEntsBank}~\cite{Dzikovska2013SemEval} provides short-answer science questions; in our experiments it includes 10,804 responses across 200 questions. \textbf{Beetle}~\cite{Dzikovska2013SemEval} contains short-answer questions about electronics and circuits with 3,559 responses across 56 questions and an average answer length of about 10 words. Each instance consists of a question, student response, rubric item, and binary correctness label.

\textbf{Models.} We evaluate seven LLMs spanning two orders of magnitude in parameter count. \textit{Small models} ($<$20B parameters) include Llama-3.1-8B-Instruct~\cite{MetaLlama3}, Qwen3-4B-Thinking~\cite{yang2025qwen3}, and GPT-OSS-20B~\cite{agarwal2025gpt}. \textit{Large models} ($\geq$70B parameters) include GPT-OSS-120B~\cite{agarwal2025gpt}, Llama-3.3-70B-Instruct~\cite{MetaLlama3}, Qwen3-Next-80B-A3B-Thinking, and Qwen3-Next-80B-A3B-Instruct~\cite{yang2025qwen3}. The Qwen3-80B variants allow comparison between standard instruction-following and extended thinking modes within the same architecture.

\textbf{Implementation Details.} All experiments use the vLLM backend with nucleus sampling~\cite{Holtzman2020,Kwon2023vLLM}. Large models run on four NVIDIA A100 GPUs with tensor parallelism, while small models execute on a bench-top GeForce RTX 3090. Sampling parameters follow model card recommendations (temperature 0.6--0.7, top-$p$ 0.8--0.95, top-$k$ 20 where applicable). We set maximum generation length to 8,192 tokens for large models and 30,000 tokens for small models, and batch sizes to 16 (large) and 128 (small). Models return JSON-formatted verdicts; self-consistency uses $n=5$ and majority vote. We report raw confidence scores throughout.

\subsection{Main Results}

Tables~\ref{tab:step2_means_small} and~\ref{tab:step2_means_big} summarize results averaged across models within each size category. We compute percentile bootstrap confidence intervals by repeatedly resampling students with replacement; typical CI widths are ECE $\approx$0.014--0.057 and Brier $\approx$0.010--0.046 (medians 0.021 and 0.019).
\input{tables/small_models}
\input{tables/large_models}

\textbf{Finding 1: Self-reported confidence achieves the best calibration.} Across all six conditions (3 datasets $\times$ 2 model sizes), self-reported confidence consistently achieves the lowest Expected Calibration Error (ECE). Averaged across datasets, self-reported confidence achieves ECE of 0.197 for small models and 0.142 for large models. The corresponding ECE values for self-consistency are 0.240/0.221 and for token probability 0.226/0.242. Overall, self-reported ECE averages 0.166, compared to 0.229 for self-consistency and 0.235 for token probability; overall Brier follows the same ordering (0.223 vs 0.241 and 0.244).
\input{tables/method_dominance}

Paired bootstrap comparisons on identical responses (Table~\ref{tab:method_dominance}) show median deltas favoring self-reported in every dataset. Median paired CIs exclude zero. Model-level tallies show self-reported beats the comparator in 6/7 models for five rows (7/7 for RiceChem ECE), with all six rows significant in 7/7 models. Across datasets, median ECE deltas range from about $-0.076$ to $-0.100$ and median Brier deltas from about $-0.021$ to $-0.046$. Self-consistency is 38\% worse despite 5$\times$ inference cost. The discrete nature of vote counts (e.g., 3/5 vs 4/5 agreement) provides coarser confidence granularity than continuous self-reported scores. Mixed-effects calibration regressions show the strongest confidence--feature interactions under self-reported confidence (Table~\ref{tab:calreg_summary}). (Here $\beta$ measures how much a 1 SD change in a feature shifts the confidence--accuracy slope; Table~\ref{tab:calreg_summary} reports median $|\beta|$, IQRs, and significance rates.) Response-level features are most frequently significant (73.3\%), while question features are rarer (12.4\%) and rubric features show no significant effects.
\input{tables/calreg_summary}

\textbf{Finding 2: Model scale boosts accuracy more than calibration.} Larger models are consistently more accurate across all three datasets, but calibration improvements are mixed and depend on the method. For self-reported confidence, the size-averaged ECE drops from 0.197 to 0.142, with Brier decreasing from 0.252 to 0.201 ($\approx$20\%). At the dataset level, SciEntsBank shows the largest gain (ECE 0.208 → 0.121; Brier 0.255 → 0.183), Beetle sees moderate gain (ECE 0.221 → 0.156; Brier 0.271 → 0.213), and RiceChem barely moves (ECE 0.162 → 0.150; Brier 0.229 → 0.209). Counterexamples arise for token probability: ECE worsens with scale in RiceChem (0.212 → 0.250) and Beetle (0.251 → 0.262). Brier shifts only slightly in those cases (RiceChem 0.243 → 0.252; Beetle 0.266 → 0.264). Across all model–method pairs, GPT-OSS-120B self-reported delivers the lowest average ECE (0.100; Brier 0.180), while the highest AUC overall occurs for Qwen3-80B-Instruct token probability on SciEntsBank (AUC 0.728).

\textbf{Finding 3: Confidence is top-skewed across methods.} Pooling self-reported and token-probability outputs, 86\% of predictions exceed 0.8; only 0.62\% fall below 0.2 (though a few model/dataset runs exceed 1\%). Self-consistency is discrete by construction: 89.5\% of outputs are 5/5 (1.0), 6.1\% are 4/5 (0.8), and 4.5\% are 3/5 (0.6). These distributions create practical confidence floors, so thresholds should be set relative to each model/method’s observed scores rather than an intuitive midpoint.

\input{tables/calibration_fig.tex}
\subsection{Calibration Visualization}

To visualize calibration behavior, Figure~\ref{fig:calibration_curves} shows calibration curves (left) and confidence histograms (right) for GPT-OSS-120B with self-reported confidence across all three datasets. SciEntsBank tracks the diagonal most closely (lowest ECE/Brier), while RiceChem and Beetle are slightly less calibrated (higher ECE/Brier). The histograms are top-skewed with most mass above 0.8 and virtually none below 0.2, so thresholds should be calibrated to observed confidence distributions rather than a fixed midpoint.

\subsection{Per-Model Analysis}
\input{tables/large_model_paper_data}

To explain the size-averaged trends, we examine model-specific patterns. Table~\ref{tab:individual} shows results for individual large models with self-reported confidence.

\textbf{Qwen3-80B: Thinking vs Instruct tradeoff.} Comparing Qwen3-80B-Thinking with Qwen3-80B-Instruct reveals a clear tradeoff between accuracy and calibration. On RiceChem, the thinking variant achieves substantially higher accuracy (77.5\% vs 71.7\%, a 5.8 percentage point gain) with better calibration (ECE 0.142 vs 0.197; Brier 0.187 vs 0.240), suggesting that extended reasoning helps with complex chemistry explanations. In contrast, on the short‑answer datasets the thinking variant’s calibration is weaker (SciEntsBank ECE/Brier 0.161/0.198 vs 0.126/0.198; Beetle 0.198/0.233 vs 0.132/0.205), which might come occasional overconfidence when longer reasoning paths miss simpler cues.

\input{tables/gpt_oss_120b_reasoning_stats}
\textbf{GPT-OSS-120B: Best calibration with stable reasoning.} Under self-reported confidence, GPT-OSS-120B achieves average ECE of 0.100, average Brier of 0.180, and average AUC of 0.668 across all three datasets, substantially outperforming other large models under self-reported confidence (avg ECE 0.157, avg Brier 0.209, avg AUC 0.615). This represents 36\% lower ECE, 14\% lower Brier, and 9\% higher AUC. Table~\ref{tab:gptoss_reasoning} shows GPT-OSS-120B performance across reasoning levels (high, medium, low). Unlike Qwen's large variation between thinking modes, GPT-OSS-120B exhibits remarkable stability: accuracy varies by at most 1.2 percentage points and ECE by at most 0.009 across reasoning levels. On SciEntsBank, all three reasoning levels achieve identical accuracy (79.1\%) with ECE ranging only from 0.081 to 0.085. This stability suggests that GPT-OSS-120B's calibration quality is robust to reasoning depth, making it a reliable choice for deployment without careful tuning of reasoning parameters.

\subsection{Practical Implications}

For deployment, well-calibrated confidence enables selective automation, where high-confidence predictions are processed automatically while uncertain cases are flagged for human review. With GPT-OSS-120B on SciEntsBank, Coverage@10\%Risk is about 20\% and Coverage@5\%Risk is 11.5\%; the highest Coverage@5\%Risk we observe is 19.5\% (Llama-3.3-70B-Instruct, self-reported, SciEntsBank). This also sets a practical limitation: even at 5\% risk, only about 11\%--20\% of predictions qualify in our best-performing runs, so selective automation covers a modest share under strict reliability targets.

%% file: tables/small_models.tex
\begin{table}[t]
\centering
\caption{Calibration metrics averaged across small models (Llama-8B, Qwen-4B, GPT-OSS-20B). Self-reported confidence achieves the lowest ECE across all datasets. Lower ECE and Brier scores indicate better calibration; higher AUC and Coverage indicate better discrimination.}
\label{tab:step2_means_small}
\begin{tabular}{llrrrrr}
\hline
Dataset & Method & Acc & ECE & Brier & AUC & Cov@10\% \\
\hline
RiceChem & self-reported & \textbf{72.4} & \textbf{0.162} & \textbf{0.229} & \textbf{0.593} & 0.005 \\
RiceChem & self-consistency & 71.9 & 0.229 & 0.251 & 0.575 & 0.000 \\
RiceChem & token-probability & \textbf{72.4} & 0.212 & 0.243 & 0.565 & \textbf{0.006} \\
SciEntsBank & self-reported & 74.9 & \textbf{0.208} & 0.255 & \textbf{0.588} & \textbf{0.094} \\
SciEntsBank & self-consistency & 74.4 & 0.227 & 0.238 & 0.561 & 0.003 \\
SciEntsBank & token-probability & \textbf{75.4} & 0.215 & \textbf{0.227} & 0.577 & 0.087 \\
Beetle & self-reported & 69.9 & \textbf{0.221} & 0.271 & 0.570 & \textbf{0.034} \\
Beetle & self-consistency & 69.4 & 0.264 & 0.277 & \textbf{0.571} & 0.001 \\
Beetle & token-probability & \textbf{71.2} & 0.251 & \textbf{0.266} & 0.557 & 0.001 \\
\hline
\end{tabular}
\end{table}

%% file: tables/large_models.tex
\begin{table}[t]
\centering
\caption{Calibration metrics averaged across large models (GPT-OSS-120B, Llama-70B, Qwen-80B variants). Self-reported confidence consistently achieves the lowest ECE, with substantial improvement over small models.}
\label{tab:step2_means_big}
\begin{tabular}{llrrrrr}
\hline
Dataset & Method & Acc & ECE & Brier & AUC & Cov@10\% \\
\hline
RiceChem & self-reported & 74.2 & \textbf{0.150} & \textbf{0.209} & 0.612 & \textbf{0.038} \\
RiceChem & self-consistency & \textbf{75.4} & 0.218 & 0.228 & 0.563 & 0.000 \\
RiceChem & token-probability & 74.4 & 0.250 & 0.252 & \textbf{0.644} & 0.000 \\
SciEntsBank & self-reported & \textbf{78.3} & \textbf{0.121} & \textbf{0.183} & 0.639 & \textbf{0.310} \\
SciEntsBank & self-consistency & \textbf{78.3} & 0.198 & 0.204 & 0.550 & 0.004 \\
SciEntsBank & token-probability & 78.1 & 0.215 & 0.216 & \textbf{0.657} & 0.198 \\
Beetle & self-reported & \textbf{73.5} & \textbf{0.156} & \textbf{0.213} & \textbf{0.635} & \textbf{0.166} \\
Beetle & self-consistency & 72.7 & 0.249 & 0.258 & 0.539 & 0.001 \\
Beetle & token-probability & 73.2 & 0.262 & 0.264 & 0.618 & 0.012 \\
\hline
\end{tabular}
\end{table}

%% file: tables/method_dominance.tex
\begin{table}[t]
\centering
\caption{Paired dominance summaries from Step~6 (model-level paired bootstrap on identical responses; within-dataset FDR). $\Delta$ $=$ self-reported $-$ comparator (negative favors self-reported). CI is the median model-level paired bootstrap interval across models. B/W/T $=$ number of models where self-reported is better/worse/tied; Sig $=$ number significant.}
\label{tab:method_dominance}
\begin{tabular}{lllrc cl}
\hline
Dataset & Metric & Comparator & $\Delta$ & CI & B/W/T; Sig \\
\hline
RiceChem & ECE & self-consistency & $-0.0860$ & $[-0.0886,-0.0703]$ & 7/0/0\; 7/7 \\
SciEntsBank & ECE & self-consistency & $-0.0753$ & $[-0.0820,-0.0684]$ & 6/1/0\; 7/7 \\
Beetle & ECE & token-probability & $-0.1000$ & $[-0.1078,-0.0910]$ & 6/1/0\; 7/7 \\
RiceChem & Brier & token-probability & $-0.0380$ & $[-0.0399,-0.0357]$ & 6/1/0\; 7/7 \\
SciEntsBank & Brier & self-consistency & $-0.0213$ & $[-0.0258,-0.0190]$ & 6/1/0\; 7/7 \\
Beetle & Brier & token-probability & $-0.0459$ & $[-0.0503,-0.0401]$ & 6/1/0\; 7/7 \\
\hline
\end{tabular}
\end{table}

%% file: tables/calreg_summary.tex
\begin{table}[t]
\centering
\small
\setlength{\tabcolsep}{4pt}
\renewcommand{\arraystretch}{0.9}
\caption{Calibration-regression interactions (confidence $\times$ feature). Rates are significant interactions (FDR within dataset). Median $|\beta|$ indicates effect size; larger values indicate stronger shifts in the confidence--accuracy slope.}
\label{tab:calreg_summary}
\begin{tabular*}{\linewidth}{@{\extracolsep{\fill}} llccc}
\hline
Dataset & Method & med$|\beta|$ & IQR [Q1--Q3] & \% sig \\
\hline
\multirow{3}{*}{RiceChem} & self-reported & \textbf{0.246} & 0.134--0.396 & \textbf{43.9} \\
& self-consistency & 0.047 & 0.036--0.068 & 27.1 \\
& token-probability & 0.075 & 0.056--0.092 & 12.4 \\
\hline
\multirow{3}{*}{SciEntsBank} & self-reported & \textbf{0.162} & 0.056--0.320 & \textbf{35.3} \\
& self-consistency & 0.075 & 0.044--0.108 & \textbf{35.3} \\
& token-probability & 0.059 & 0.043--0.109 & 14.9 \\
\hline
\multirow{3}{*}{Beetle} & self-reported & \textbf{0.155} & 0.091--0.250 & \textbf{23.9} \\
& self-consistency & 0.068 & 0.058--0.083 & 15.8 \\
& token-probability & 0.076 & 0.061--0.077 & 9.9 \\
\hline
\end{tabular*}
\vspace{0.2em}
\footnotesize\parbox{\linewidth}{All coefficients are standardized.}
\end{table}

%% file: tables/calibration_fig.tex
\begin{figure}[!ht]
\centering
\begin{subfigure}[b]{0.48\textwidth}
    \includegraphics[width=\textwidth]{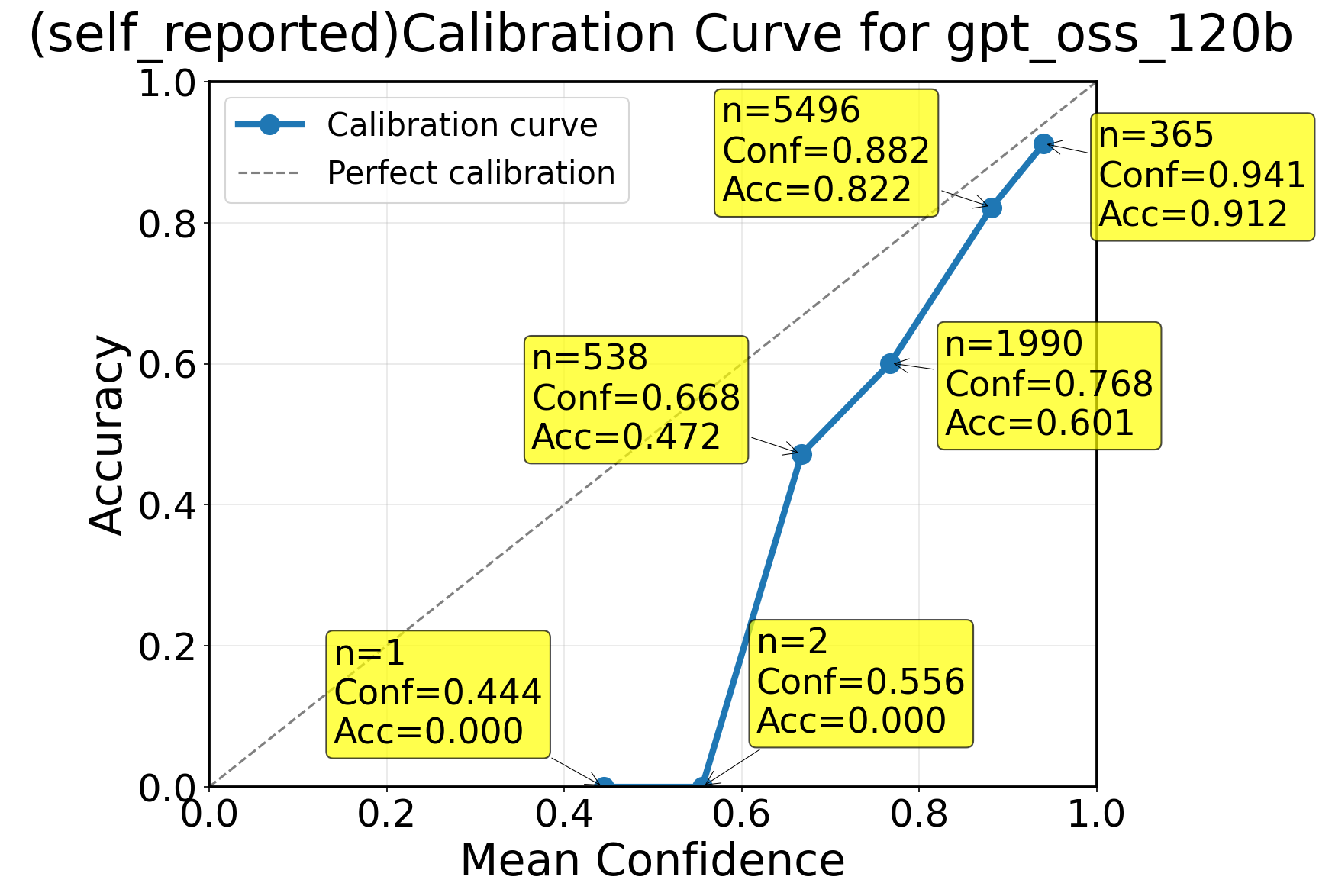}
    \caption{RiceChem: calibration curve}
\end{subfigure}
\hfill
\begin{subfigure}[b]{0.48\textwidth}
    \includegraphics[width=\textwidth]{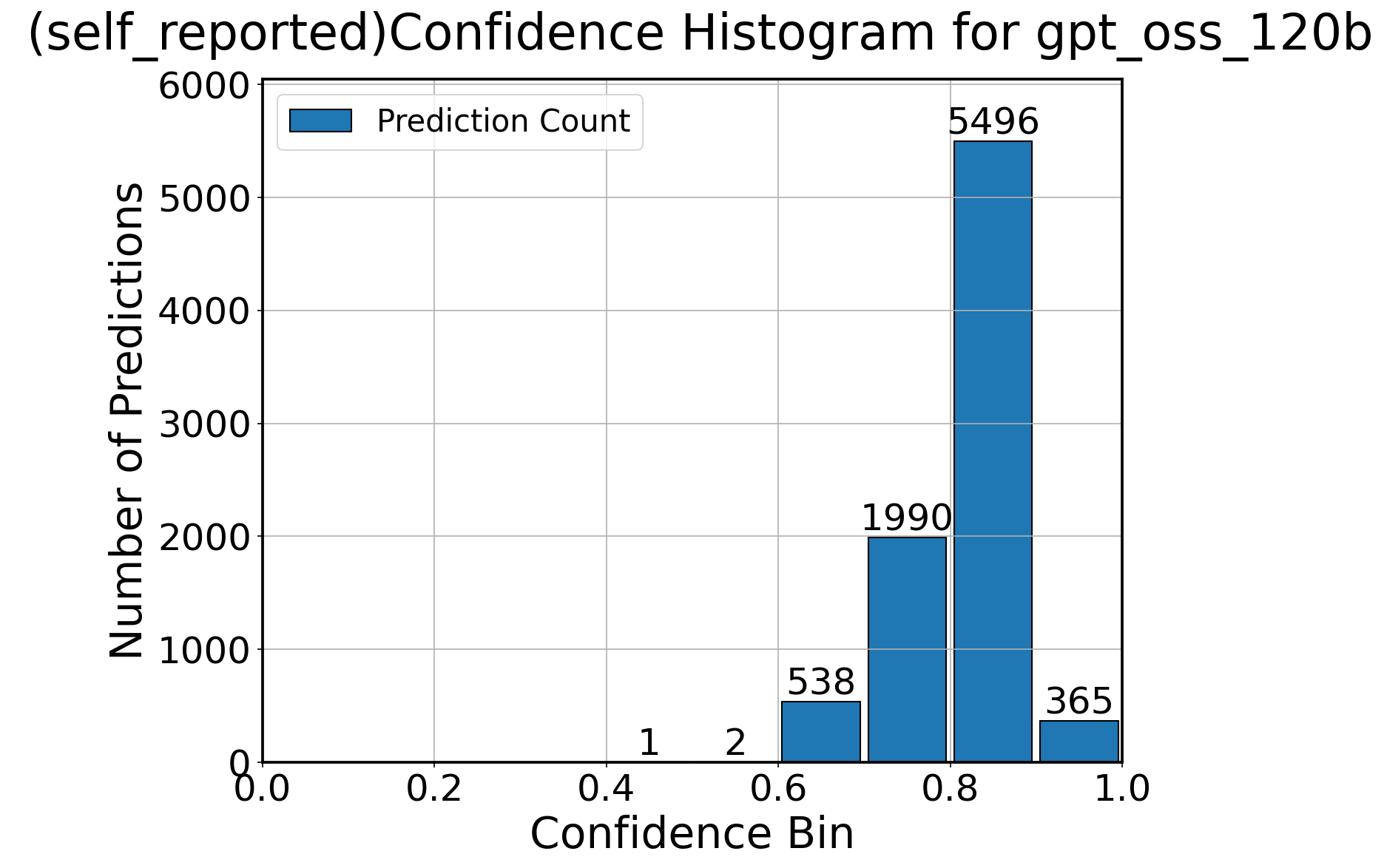}
    \caption{RiceChem: confidence distribution}
\end{subfigure}
\\[0.3em]
\begin{subfigure}[b]{0.48\textwidth}
    \includegraphics[width=\textwidth]{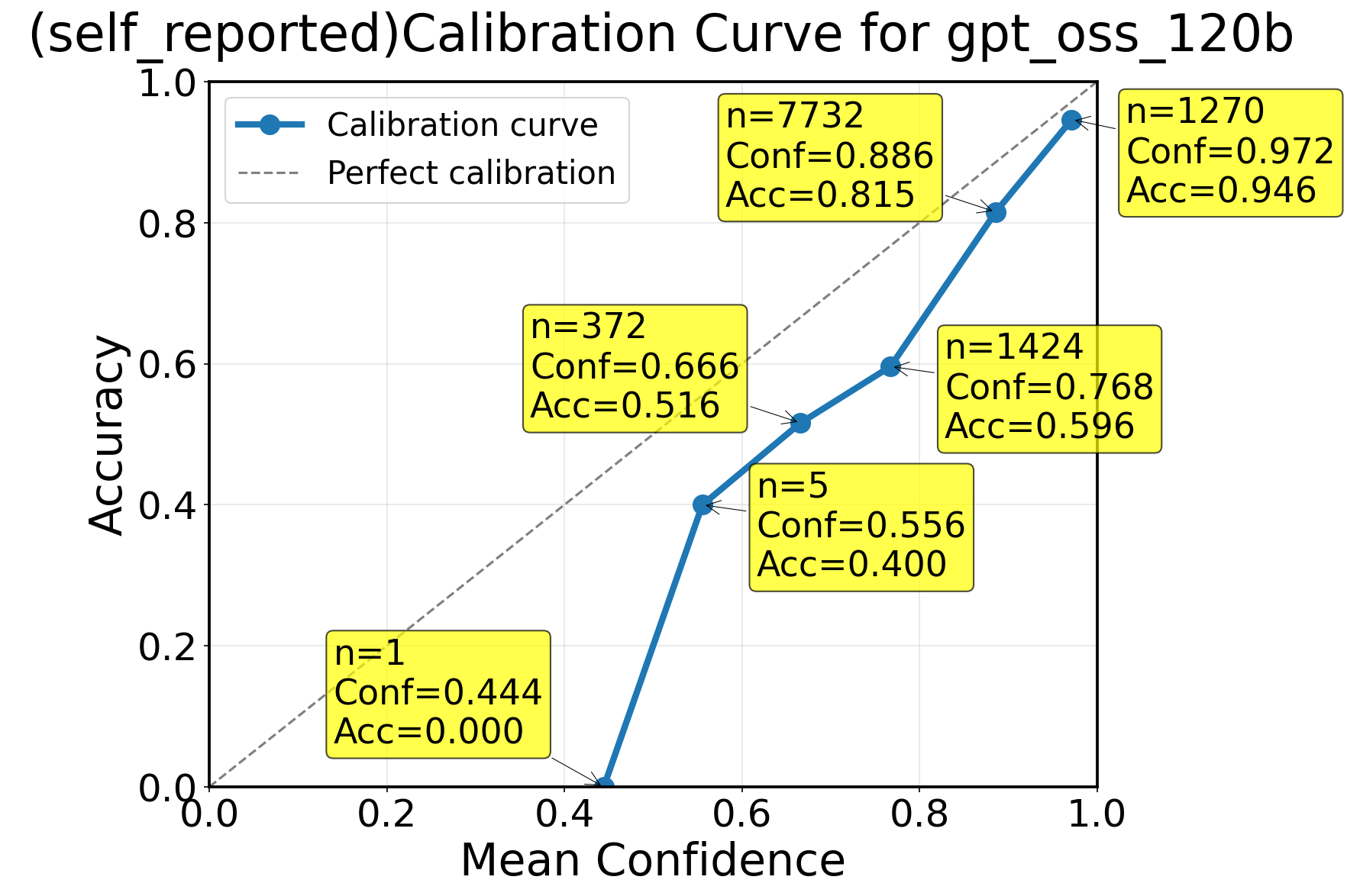}
    \caption{SciEntsBank: calibration curve}
\end{subfigure}
\hfill
\begin{subfigure}[b]{0.48\textwidth}
    \includegraphics[width=\textwidth]{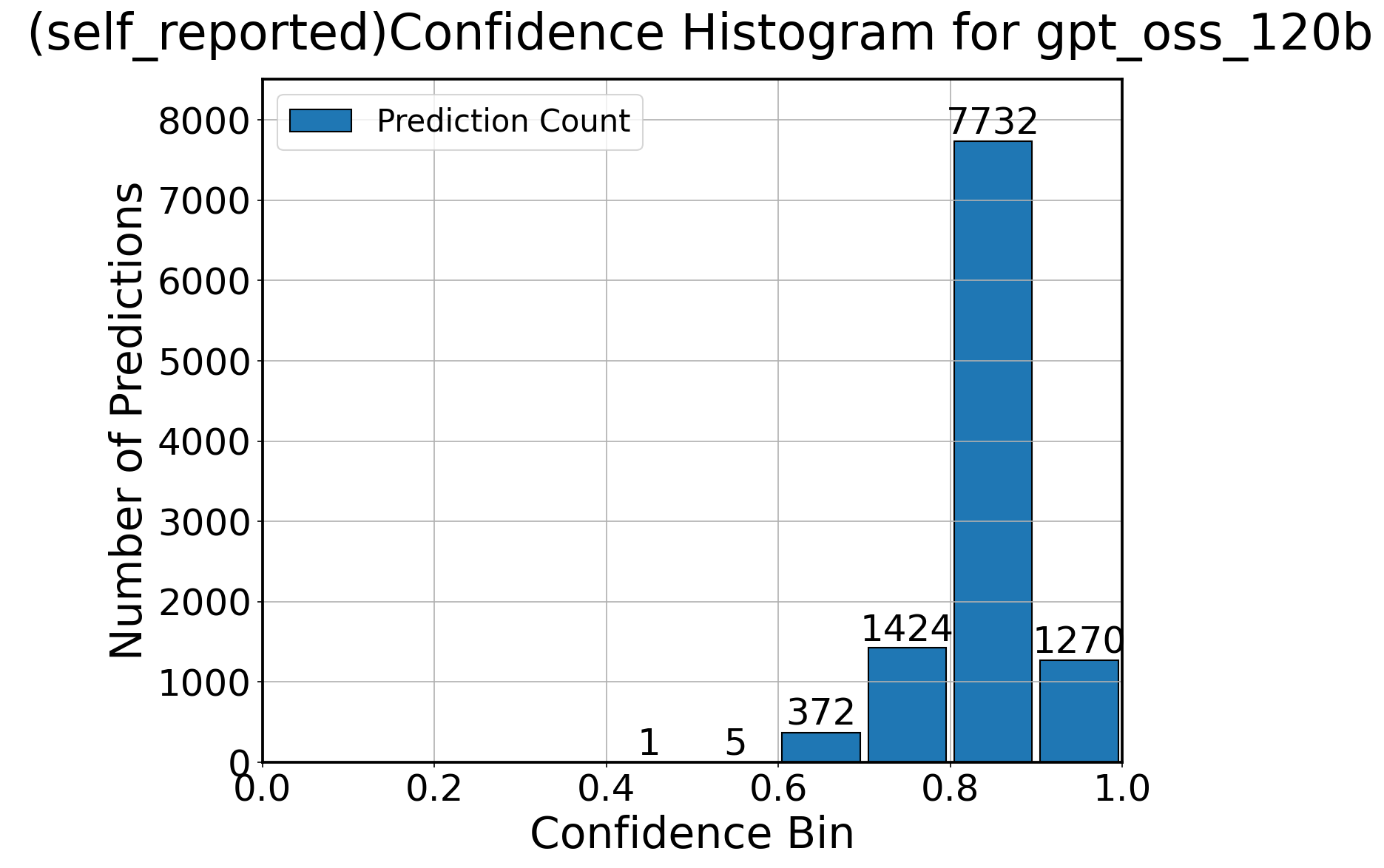}
    \caption{SciEntsBank: confidence distribution}
\end{subfigure}
\\[0.3em]
\begin{subfigure}[b]{0.48\textwidth}
    \includegraphics[width=\textwidth]{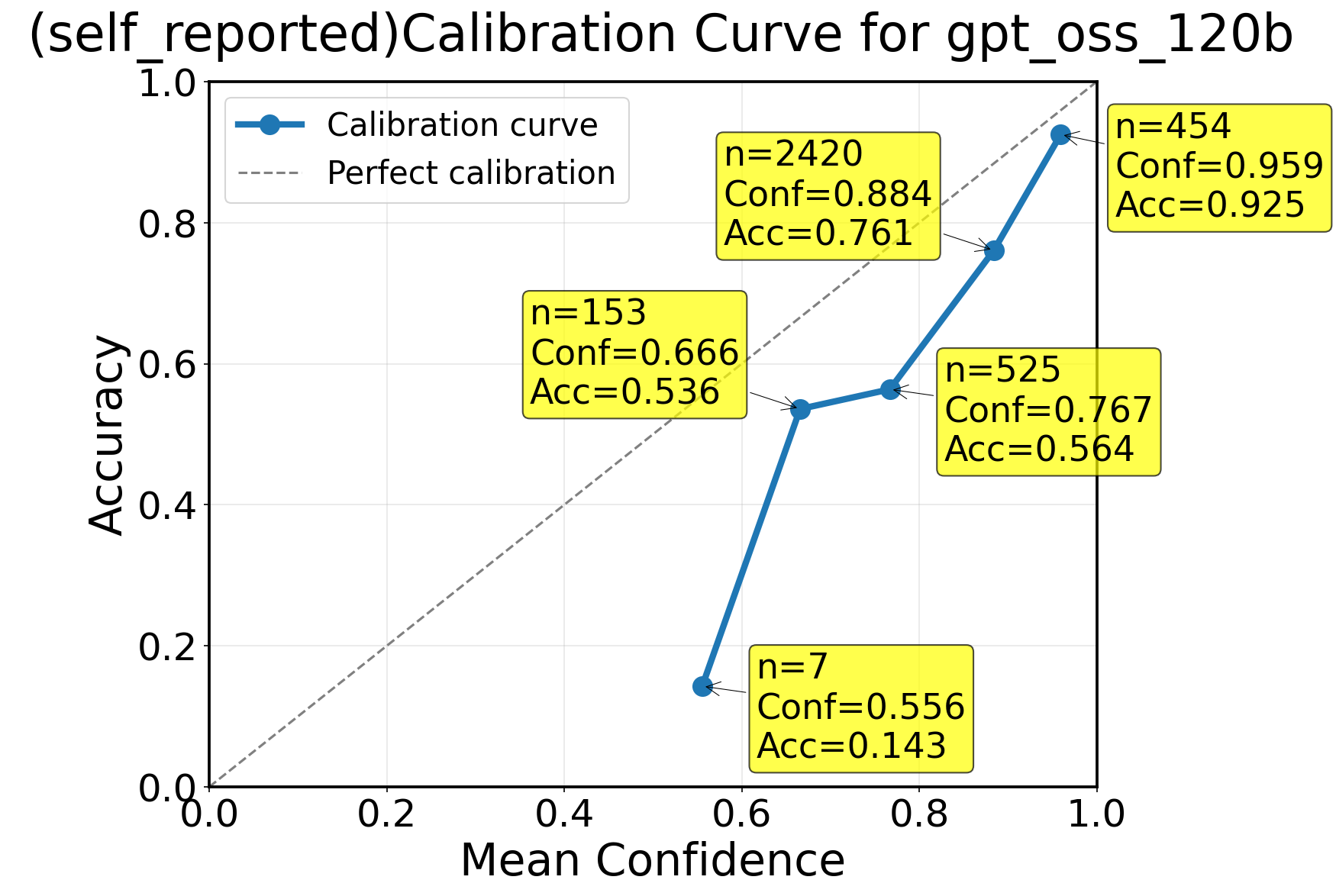}
    \caption{Beetle: calibration curve}
\end{subfigure}
\hfill
\begin{subfigure}[b]{0.48\textwidth}
    \includegraphics[width=\textwidth]{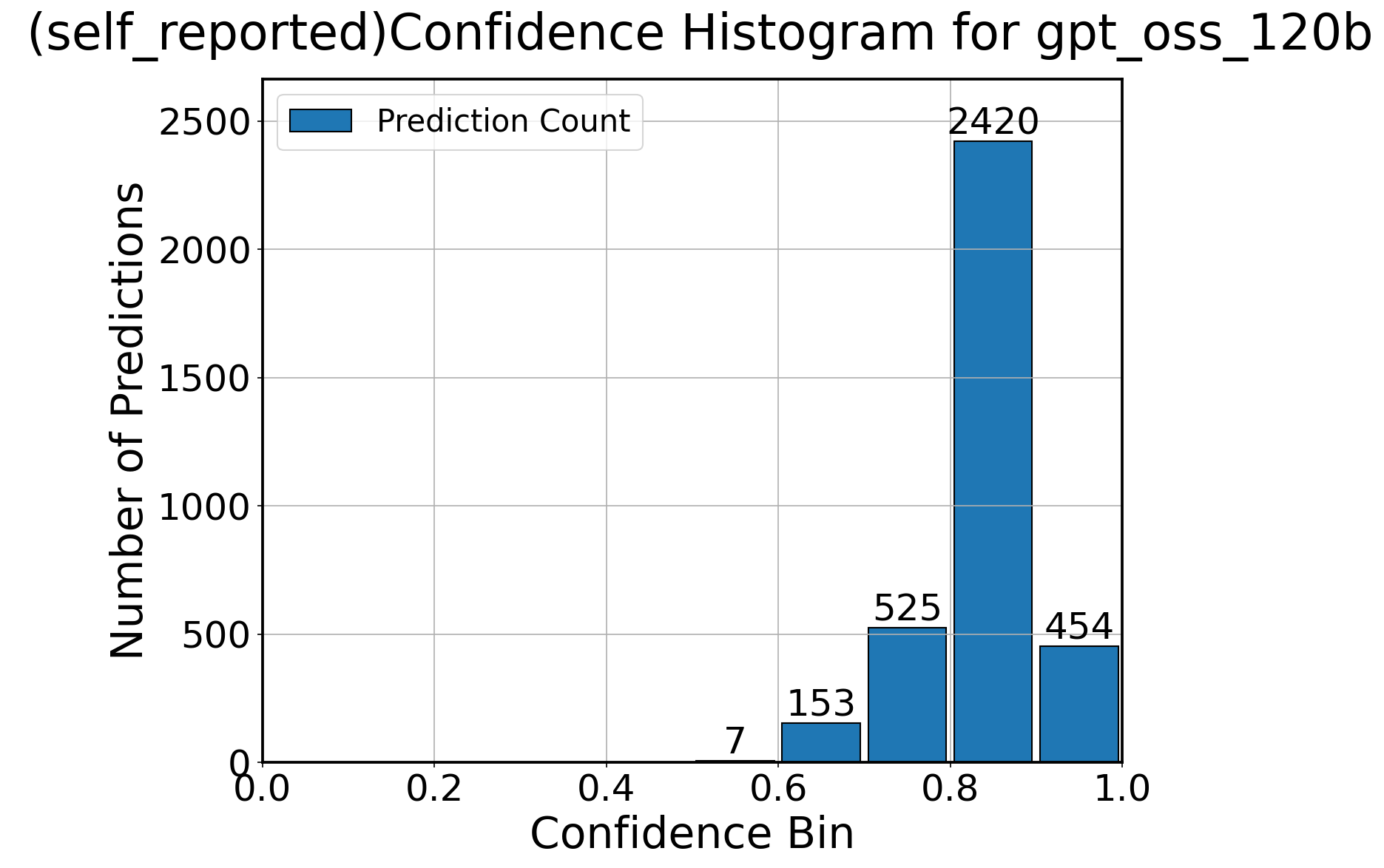}
    \caption{Beetle: confidence distribution}
\end{subfigure}
\caption{Calibration curves and confidence distributions for GPT-OSS-120B with self-reported confidence. Left column: calibration curves where the diagonal indicates perfect calibration and points below the diagonal indicate overconfidence. Right column: histograms showing that confidence scores cluster in the 0.7--0.9 range, with fewer than 1\% of predictions below 0.6.}
\label{fig:calibration_curves}
\end{figure}

%% file: tables/large_model_paper_data.tex
\begin{table}[t]
\centering
\caption{Individual large model results with self-reported confidence. ECE = Expected Calibration Error, Brier = Brier Score, AUC = AUC-ROC for correctness prediction. Best values per dataset shown in bold.}
\label{tab:individual}
\begin{tabular}{lrrrrrrrrrrrr}
\hline
Model & \multicolumn{4}{c}{RiceChem} & \multicolumn{4}{c}{SciEntsBank} & \multicolumn{4}{c}{Beetle} \\
& Acc & ECE & Brier & AUC & Acc & ECE & Brier & AUC & Acc & ECE & Brier & AUC \\
\hline
GPT-OSS-120B & 75.1 & \textbf{.093} & \textbf{.182} & \textbf{.690} & \textbf{79.1} & \textbf{.081} & \textbf{.162} & \textbf{.656} & 74.2 & \textbf{.125} & \textbf{.196} & \textbf{.660} \\
Llama-3.3-70B & 72.4 & .166 & .226 & .527 & 79.0 & .117 & .173 & .602 & 73.3 & .169 & .218 & .620 \\
Qwen3-80B-Think & \textbf{77.5} & .142 & .187 & .636 & 76.4 & .161 & .198 & .645 & 72.1 & .198 & .233 & .618 \\
Qwen3-80B-Inst & 71.7 & .197 & .240 & .594 & 78.7 & .126 & .198 & .655 & \textbf{74.5} & .132 & .205 & .641 \\
\hline
\end{tabular}
\end{table}

%% file: tables/gpt_oss_120b_reasoning_stats.tex
\begin{table}[t]
\centering
\caption{GPT-OSS-120B performance across reasoning levels with self-reported confidence. Medium is the default setting used in main experiments. Best values per dataset shown in bold.}
\label{tab:gptoss_reasoning}
\begin{tabular}{lrrrrrrrrrrrr}
\hline
Reasoning & \multicolumn{4}{c}{RiceChem} & \multicolumn{4}{c}{SciEntsBank} & \multicolumn{4}{c}{Beetle} \\
& Acc & ECE & Brier & AUC & Acc & ECE & Brier & AUC & Acc & ECE & Brier & AUC \\
\hline
High & 74.5 & .099 & .186 & .690 & 79.1 & .085 & .162 & \textbf{.663} & \textbf{75.4} & \textbf{.116} & \textbf{.186} & \textbf{.674} \\
Medium & \textbf{75.1} & \textbf{.093} & \textbf{.182} & .690 & \textbf{79.1} & \textbf{.081} & \textbf{.162} & .656 & 74.2 & .125 & .196 & .660 \\
Low & 74.3 & .101 & .186 & \textbf{.698} & 79.1 & .085 & .162 & .660 & 75.0 & .119 & .189 & .671 \\
\hline
\end{tabular}
\end{table}

%% file: sections/conclusion.tex
\section{Conclusion}

We evaluated which confidence signal can best tell us when an LLM grader is reliable, comparing self‑consistency, self‑reported confidence, and token‑probability estimates across seven models and three educational datasets. This work keeps experimental conditions (prompts, rubrics, student populations) consistent so the comparisons are directly comparable and the calibration statistics are interpretable.
Finding 1: Self‑reported confidence yields the most calibrated signal within these ASAG settings, outperforming sampling‑based alternatives while keeping inference cost low.
Finding 2: Larger models deliver steadier accuracy gains, but calibration improvements remain dataset‑ and method‑dependent.
Finding 3: Confidence scores are top‑skewed, so thresholds should be set relative to each method’s observed distribution rather than an intuitive midpoint.
Taken together, these results support using self‑reported confidence as the automation gate while recognizing that strict reliability targets cover only a modest share of predictions. Mixed‑model and paired‑bootstrap analyses provide the statistical backing that makes this practical guidance trustworthy.

\begingroup
\setlength{\itemsep}{0pt}
\setlength{\parskip}{0pt}
\bibliographystyle{splncs04}
\bibliography{refs}
\endgroup

%% file: refs.bib
@InProceedings{Guo2017ICML,
  title = 	 {On Calibration of Modern Neural Networks},
  author =       {Guo, Chuan and Pleiss, Geoff and Sun, Yu and Weinberger, Kilian Q.},
  publisher = {JMLR.org},
  booktitle = {Proc. ICML},
  pages = 	 {1321--1330},
  year = 	 {2017}
}

@inproceedings{Holtzman2020,
  title        = {The Curious Case of Neural Text Degeneration},
  author       = {Holtzman, Ari and Buys, Jan and Du, Li and Forbes, Maxwell and Choi, Yejin},
  booktitle    = {Proc. ICLR},
  year         = {2020}
}

@inproceedings{Hendrycks2017,
  author       = {Hendrycks, Dan and Gimpel, Kevin},
  title        = {A Baseline for Detecting Misclassified and Out-of-Distribution Examples
                  in Neural Networks},
  booktitle    = {Proc. ICLR},
  year         = {2017}
}

@book{EfronTibshirani1993,
  title        = {An Introduction to the Bootstrap},
  author       = {Efron, Bradley and Tibshirani, Robert J.},
  publisher    = {Chapman \& Hall/CRC},
  address      = {New York},
  year         = {1994}
}

@article{FieldWelsh2007,
  title        = {Bootstrapping Clustered Data},
  author       = {Field, C. A. and Welsh, A. H.},
  journal      = {Journal of the Royal Statistical Society: Series B (Statistical Methodology)},
  year         = {2007},
  volume       = {69},
  number       = {3},
  pages        = {369--390},
  doi          = {10.1111/j.1467-9868.2007.00593.x}
}

@book{fielding2006cross,
  title={Cross-classified and multiple membership structures in multilevel models: An introduction and review},
  author={Fielding, Antony and Goldstein, Harvey},
  volume={791},
  year={2006},
  publisher={Department for Education and Skills London}
}

@article{Noortgate2003,
  title        = {Cross-Classification Multilevel Logistic Models in Psychometrics},
  author       = {Van den Noortgate, Wim and De Boeck, Paul and Meulders, Michel},
  journal      = {Journal of Educational and Behavioral Statistics},
  year         = {2003},
  doi          = {10.3102/10769986028004369}
}

@article{Dalton2013,
  title        = {Flexible Recalibration of Binary Clinical Prediction Models},
  author       = {Dalton, J. E.},
  journal      = {Statistics in Medicine},
  year         = {2013},
  volume       = {32},
  number       = {2},
  pages        = {282--289},
  doi          = {10.1002/sim.5544}
}

@inproceedings{Koehn2004,
  title        = {Statistical Significance Tests for Machine Translation Evaluation},
  author       = {Koehn, Philipp},
  booktitle    = {Proceedings of the 2004 Conference on Empirical Methods in Natural Language Processing (EMNLP)},
  year         = {2004},
  organization = {Association for Computational Linguistics}
}

@article{Brier1950,
  title        = {Verification of Forecasts Expressed in Terms of Probability},
  author       = {Brier, Glenn W.},
  journal      = {Monthly Weather Review},
  volume       = {78},
  number       = {1},
  pages        = {1--3},
  year         = {1950},
  doi          = {10.1175/1520-0493(1950)078<0001:VOFEIT>2.0.CO;2}
}

@article{Fawcett2006,
  title        = {An Introduction to {ROC} Analysis},
  author       = {Fawcett, Tom},
  journal      = {Pattern Recognition Letters},
  volume       = {27},
  number       = {8},
  pages        = {861--874},
  year         = {2006},
  doi          = {10.1016/j.patrec.2005.10.010}
}

@inproceedings{Kwon2023vLLM,
  title        = {{vLLM}: Easy, Fast, and Cheap {LLM} Serving with {PagedAttention}},
  author       = {Kwon, Woosuk and Li, Zhuohan and Zhuang, Yuxin and Sheng, Yuntian and Zhao, Nianjun and Zhang, Hao and Gonzalez, Joseph and Stoica, Ion},
  journal      = {arXiv preprint arXiv:2309.06180},
  year         = {2023}
,
  title={Efficient Memory Management for Large Language Model Serving with PagedAttention},
  author={Woosuk Kwon and Zhuohan Li and Siyuan Zhuang and Ying Sheng and Lianmin Zheng and Cody Hao Yu and Joseph E. Gonzalez and Hao Zhang and Ion Stoica},
  booktitle={Proceedings of the ACM SIGOPS 29th Symposium on Operating Systems Principles},
  year={2023}
}

@incollection{Platt1999,
  title        = {Probabilistic Outputs for Support Vector Machines and Comparisons to Regularized Likelihood Methods},
  author       = {Platt, John C.},
  booktitle    = {Advances in Large Margin Classifiers},
  pages        = {61--74},
  publisher={Cambridge, MA},
  year         = {1999}
}

@inproceedings{ZadroznyElkan2002,
  title        = {Transforming Classifier Scores into Accurate Multiclass Probability Estimates},
  author       = {Zadrozny, Bianca and Elkan, Charles},
  booktitle    = {Proceedings of the Eighth {ACM} {SIGKDD} International Conference on Knowledge Discovery and Data Mining.},
  year         = {2002},
  publisher    = {ACM}
}

@inproceedings{Dzikovska2012,
    title = "Towards Effective Tutorial Feedback for Explanation Questions: A Dataset and Baselines",
    author = "Dzikovska, Myroslava O. and Nielsen, Rodney D. and Brew, Chris",
    booktitle = "Proc. NAACL-HLT",
    year = "2012",
    pages = "200--210"
}

@inproceedings{Dzikovska2013SemEval,
  title = {{S}em{E}val-2013 Task 7: The Joint Student Response Analysis and 8th Recognizing Textual Entailment Challenge},
  author = {Dzikovska, Myroslava and Nielsen, Rodney and Brew, Chris and Leacock, Claudia and Giampiccolo, Danilo and Bentivogli, Luisa and Clark, Peter and Dagan, Ido and Dang, Hoa Trang},
  year = 2013,
  booktitle = {Proc. SemEval},
  pages = {263--274}
}

@InProceedings{RiceChem2024,
author={Sonkar, Shashank and Ni, Kangqi and Tran Lu, Lesa and Kincaid, Kristi and Hutchinson, John S. and Baraniuk, Richard G.},
title={Automated Long Answer Grading with RiceChem Dataset},
booktitle={Artificial Intelligence in Education},
year={2024},
pages={163--176}
}

@article{Kostic2024, title={LLMs in Automated Essay Evaluation: A Case Study}, volume={3}, DOI={10.1609/aaaiss.v3i1.31193}, abstractNote={This study delves into the application of large language models (LLMs), such as ChatGPT-4, for the automated evaluation of student essays, with a focus on a case study conducted at the Swiss Institute of Business Administration. It explores the effectiveness of LLMs in assessing German-language student transfer assignments, and contrasts their performance with traditional evaluations by human lecturers. The primary findings highlight the challenges faced by LLMs in terms of accurately grading complex texts according to predefined categories and providing detailed feedback. This research illuminates the gap between the capabilities of LLMs and the nuanced requirements of student essay evaluation. The conclusion emphasizes the necessity for ongoing research and development in the area of LLM technology to improve the accuracy, reliability, and consistency of automated essay assessments in educational contexts.}, number={1}, journal={Proceedings of the AAAI Symposium Series}, author={Kostic, Milan and Witschel, Hans Friedrich and Hinkelmann, Knut and Spahic-Bogdanovic, Maja}, year={2024}, month={May}, pages={143-147} }

@article{Gandolfi2025,
  title   = {{GPT}-4 in Education: Evaluating Aptness, Reliability, and Loss of Coherence in Solving Calculus Problems and Grading Submissions},
  author  = {Gandolfi, Alberto},
  journal = {International Journal of Artificial Intelligence in Education},
  year    = {2025},
  publisher = {Springer},
  DOI={https://doi.org/10.1007/s40593-024-00403-3}
}

@inproceedings{Tian2024Artifacts,
 author = {Xiaoyi Tian and Amogh Mannekote and Others},
 booktitle = {Proceedings of the 17th International Conference on Educational Data Mining},
 doi = {10.5281/zenodo.12729922},
 editor = {Benjamin PaaÃŸen and Carrie Demmans Epp},
 isbn = {978-1-7336736-5-5},
 month = {July},
 pages = {698--706},
 title = {Examining LLM Prompting Strategies for Automatic Evaluation of Learner-Created Computational Artifacts},
 year = {2024}
}

@article{Usher2025PeerGrading,
author = {Maya Usher and Montathar Faraon},
title = {Who grades best? Comparing ChatGPT, peer, and instructor evaluations across varying levels of student project quality},
journal = {Assessment \& Evaluation in Higher Education},
volume = {0},
number = {0},
pages = {1--20},
year = {2025},
publisher = {Routledge},
doi = {10.1080/02602938.2025.2588682},
eprint = {
    https://doi.org/10.1080/02602938.2025.2588682
}

}

@inproceedings{
Wang2023SelfConsist,
title={Self-Consistency Improves Chain of Thought Reasoning in Language Models},
author={Wang, Xuezhi and Wei, Jason and Schuurmans, Dale and Le, Quoc V. and Chi, Ed H. and Narang, Sharan and Chowdhery, Aakanksha and Zhou, Denny},
booktitle={Proc. ICLR},
year={2023}
}

@article{Kadavath2022,
  title   = {Language Models (Mostly) Know What They Know},
  author  = {Kadavath, Saurav and Conerly, Tom and Askell, Amanda and Henighan, Tom and Drain, Dawn and Perez, Ethan and Schiefer, Nicholas and Hatfield-Dodds, Zac and DasSarma, Nova and Tran-Johnson, Eli and others},
  journal = {arXiv preprint arXiv:2207.05221},
  year    = {2022},
  note    = {Anthropic}
}

@inproceedings{Kuhn2023ICLR,
  title     = {Semantic Uncertainty: Linguistic Invariances for Uncertainty Estimation in Natural Language Generation},
  author    = {Kuhn, Lorenz and Gal, Yarin and Farquhar, Sebastian},
  booktitle = {International Conference on Learning Representations (ICLR)},
  year      = {2023}
}

@inproceedings{Xiong2024ICLR,
  title     = {Can {LLMs} Express Their Uncertainty? An Empirical Evaluation of Confidence Elicitation in {LLMs}},
  author    = {Xiong, Miao and Hu, Zhiyuan and Lu, Xinyang and Li, Yifei and Fu, Jie and He, Junxian and Hooi, Bryan},
  booktitle = {International Conference on Learning Representations (ICLR)},
  year      = {2024}
}

@inproceedings{Geng2024NAACL,
  title     = {A Survey of Confidence Estimation and Calibration in Large Language Models},
  author    = {Geng, Jiahui  and
      Cai, Fengyu  and
      Wang, Yuxia  and
      Koeppl, Heinz  and
      Nakov, Preslav  and
      Gurevych, Iryna},
  booktitle = {Proceedings of the 2024 Conference of the North American Chapter of the Association for Computational Linguistics (NAACL)},
  year      = {2024}
}

@inproceedings{Tian2023EMNLP,
  title     = {Just Ask for Calibration: Strategies for Eliciting Calibrated Confidence Scores from Language Models Fine-Tuned with Human Feedback},
  author    = {Tian, Katherine and Mitchell, Eric and Zhou, Allan and Sharma, Archit and Rafailov, Rafael and Yao, Huaxiu and Finn, Chelsea and Manning, Christopher D.},
  booktitle = {Proceedings of the 2023 Conference on Empirical Methods in Natural Language Processing (EMNLP)},
  year      = {2023}
}

@inproceedings{Lakshminarayanan2017NeurIPS,
  title     = {Simple and Scalable Predictive Uncertainty Estimation using Deep Ensembles},
  author    = {Lakshminarayanan, Balaji and Pritzel, Alexander and Blundell, Charles},
  booktitle = {Neural Information Processing Systems (NeurIPS)},
  year      = {2017}
}

@inproceedings{Gal2016ICML,
  title     = {Dropout as a {Bayesian} Approximation: Representing Model Uncertainty in Deep Learning},
  author    = {Gal, Yarin and Ghahramani, Zoubin},
  booktitle = {International Conference on Machine Learning (ICML)},
  year      = {2016}
}

@article{Burrows2015IJAIED,
  title   = {The Eras and Trends of Automatic Short Answer Grading},
  author  = {Burrows, Steven and Gurevych, Iryna and Stein, Benno},
  journal = {International Journal of Artificial Intelligence in Education},
  volume  = {25},
  pages   = {60--117},
  year    = {2015},
  publisher = {Springer}
}

@inproceedings{Jiang2024LAS,
  title     = {Short Answer Scoring with {GPT}-4},
  author    = {Jiang, Lan and Bosch, Nigel},
  booktitle = {Proceedings of the 11th ACM Conference on Learning @ Scale},
  pages     = {438--442},
  year      = {2024},
  publisher = {ACM}
}

@article{Pakdaman_2015, title={Obtaining Well Calibrated Probabilities Using Bayesian Binning}, volume={29},  DOI={10.1609/aaai.v29i1.9602}, journal={Proceedings of the AAAI Conference on Artificial Intelligence}, author={Pakdaman Naeini, Mahdi and Cooper, Gregory and Hauskrecht, Milos}, year={2015}}

@article{MetaLlama3,
  author       = {Llama Team},
  title        = {The Llama 3 Herd of Models},
  journal      = {CoRR},
  volume       = {abs/2407.21783},
  year         = {2024},
  doi          = {10.48550/ARXIV.2407.21783},
  eprinttype    = {arXiv},
  eprint       = {2407.21783},
  bibsource    = {dblp computer science bibliography, https://dblp.org}
}

@article{agarwal2025gpt,
  title   = {{GPT-OSS-120B} \& {GPT-OSS-20B} Model Card},
  author  = {{OpenAI}},
  journal = {arXiv preprint arXiv:2508.10925},
  year    = {2025}
}

@misc{yang2025qwen3,
      title={Qwen3 Technical Report}, 
      author={Qwen Team},
      year={2025},
      eprint={2505.09388},
      archivePrefix={arXiv},
      primaryClass={cs.CL},
      url={https://arxiv.org/abs/2505.09388}, 
}

@article{FDR_BH,
 ISSN = {00359246},
 author = {Yoav Benjamini and Yosef Hochberg},
 journal = {Journal of the Royal Statistical Society. Series B (Methodological)},
 number = {1},
 pages = {289--300},
 publisher = {[Royal Statistical Society, Oxford University Press]},
 title = {Controlling the False Discovery Rate: A Practical and Powerful Approach to Multiple Testing},
 urldate = {2026-02-02},
 volume = {57},
 year = {1995}
}
